\documentclass[10pt,twocolumn,letterpaper]{article}

\usepackage{cvpr}
\usepackage{color}
\usepackage{times}
\usepackage{epsfig}
\usepackage{graphicx}
\usepackage{amsmath}
\usepackage{amssymb}

\usepackage[breaklinks=true,bookmarks=false]{hyperref}

\cvprfinalcopy 

\def\cvprPaperID{****} 

\begin{document}

\title{Feature Evaluation of Deep Convolutional Neural Networks for Object Recognition and Detection}

\author{Hirokatsu Kataoka, Kenji Iwata, Yutaka Satoh\\
National Institute of Advanced Industrial Science and Technology (AIST)\\
Tsukuba, Ibaraki, Japan\\
{\tt\small hirokatsu.kataoka@aist.go.jp, http://www.hirokatsukataoka.net/}
}

\maketitle

\begin{abstract}
In this paper, we evaluate convolutional neural network (CNN) features using the AlexNet architecture developed by~\cite{Krizhevsky2012} and very deep convolutional network (VGGNet) architecture developed by~\cite{Simonyan2015}. To date, most CNN researchers have employed the last layers before output, which were extracted from the fully connected feature layers. However, since it is unlikely that feature representation effectiveness is dependent on the problem, this study evaluates additional convolutional layers that are adjacent to fully connected layers, in addition to executing simple tuning for feature concatenation (e.g., layer 3 + layer 5 + layer 7) and transformation, using tools such as principal component analysis. In our experiments, we carried out detection and classification tasks using the Caltech 101 and Daimler Pedestrian Benchmark Datasets. 
\end{abstract}

\section{Introduction}
Over the past few years, convolutional neural networks (CNNs) have significantly improved from the standpoint of the network architectures needed to facilitate recognition accuracy and to reduce processing costs~\cite{Schmidhuber2015}. Currently, CNNs are primarily used to help users understand objects and scenes in an image. In our study, we applied a CNN to an ImageNet dataset containing over 1.4 million images and 1,000 object categories~\cite{Russakovsky2014}. Use of such a large-scale dataset allows us to model a wide variety of object recognition image features. By using the pre-trained ImageNet dataset model, we found that CNN is capable of presenting significantly more effective feature variations. 

For feature extraction, Donahue \textit{et al.} employed CNN features as a feature vector by combining those features with a support vector machine (SVM) classifier~\cite{Donahue2014}, while other researchers have evaluated and visualized CNN features with an eight-layer AlexNet architecture~\cite{Krizhevsky2012}. More recent architectures utilize deep structures, such as the very deep convolutional network (VGGNet)~\cite{Simonyan2015} and GoogLeNet~\cite{Szegedy2015}, which were developed by Oxford University's Visual Geometry Group and Google Inc., respectively. 

According to He \textit{et al.}~\cite{HeCVPR2015}, the most important CNN feature is deep architecture. Along this line, the VGGNet contains 16 to 19 layers and GoogLeNet utilizes 22 layers. VGGNet is frequently used in the computer vision field, not only in full scratch neural net models, but also as a feature generator. CNN's utility as a feature generator is also important because it can function well even if only a few learning samples are available. Thus, large-scale databases such as ImageNet can provide recognition rates that outperform human-level classification (e.g., ~\cite{Hearxiv2015, Ioffe2015}). However, this performance will fluctuate depending on the amount and variance of the data. Therefore, when CNN is used for feature generation, it provides better performance for some recognition problems than others. 

Donahue \textit{et al.} argued that usage should be limited to the last two layers before output, which are extracted from first and second fully connected layers in CNN features with AlexNet. However, we believe that more detailed evaluations should be undertaken since several different architectures have recently been proposed, and because middle layers have not been examined as feature descriptors. Accordingly, in this study, we performed more detailed experiments to evaluate two famous CNN architectures -- AlexNet and VGGNet. In addition, we carried out simple tuning for feature concatenation (e.g., layer 3 + layer 5 + layer 7) and transformations (e.g., principal component analysis: PCA).

The rest of this paper is organized as follows. In Section 2, related works are listed. The feature settings are evaluated in Section 3. The results are shown in Section 4. Finally, we conclude the paper in Section 5.

\begin{figure*}[t]
\begin{center}
   \includegraphics[width=1.0\linewidth]{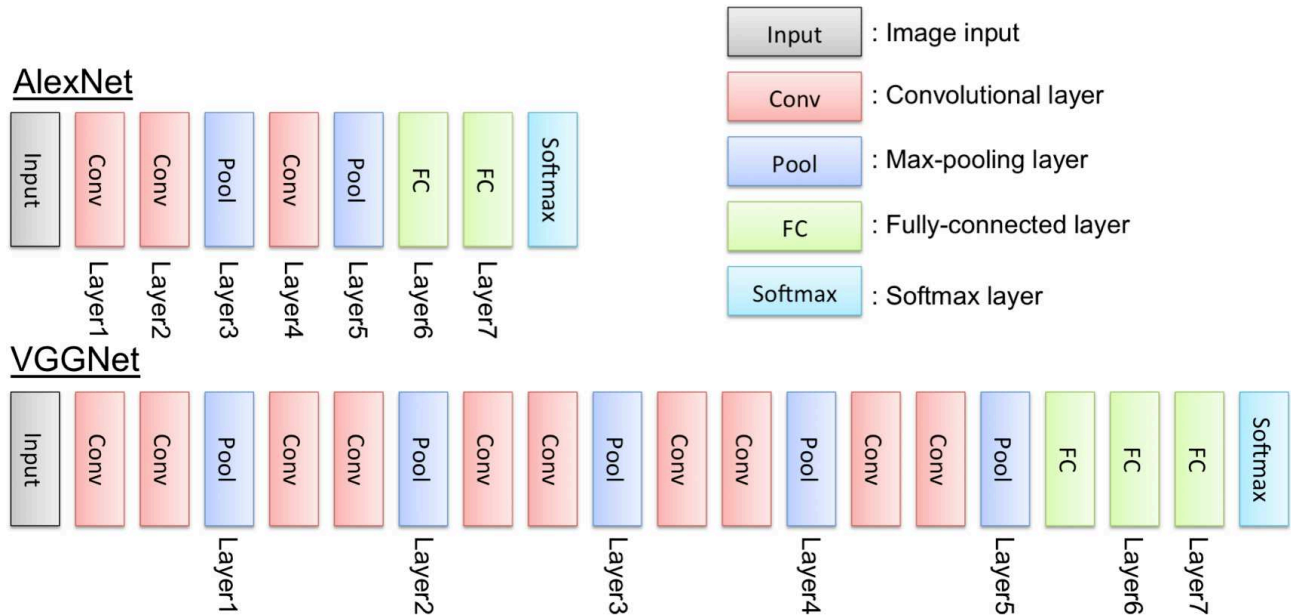}
\end{center}
   \caption{AlexNet and VGGNet architecture.}
\label{fig:alex_vgg}
\end{figure*}

\section{Related works}
In the time since the neocognitron was first proposed by Fukushima~\cite{Fukushima1980}, neuron-based recognition has become one of the most commonly used neural network architectures. Following that study, the LeNet-5~\cite{LeCun1989} neocognitron model added a baseline to CNNs in order to create a more significant model. Current network architectures include standard structures such as multiple fully connected layers, while recent challengers employ pre-trained ~\cite{Hinton2006}, dropout~\cite{Hinton2012}, and rectified linear units (ReLU)~\cite{Nair2010} as improved learning models. The most outstanding computer vision result was obtained by AlexNet in the 2012 ImageNet Large Scale Visual Recognition Challenge (ILSVRC2012), which remains the image recognition leader, with 1,000 classes~\cite{Krizhevsky2012}. 

AlexNet made it possible to increase the number of layers in network architectures. For example, Krizhevsky \textit{et al.} implemented an eight-layer model that includes convolution, pooling, and fully connected layers. More recent variations, such as the 16- or 19-layer VGGNet ~\cite{Simonyan2015}, and the 22-layer GoogLeNet~\cite{Szegedy2015} models, have even deeper architectures. These deeper models outperform conventional models on the ILSVRC dataset~\cite{Russakovsky2014}. More specifically, when compared to the AlexNet (top-five error rate on the ILSVRC2012: 16.4\%), deeper models achieved better performance levels with GoogLeNet and VGGNet (top-five error rate on the ILSVRC2014: 6.7\% for GoogLeNet and 7.3\% for VGGNet). 
Currently, the object detection problem is one of the most important topics in computer vision. The existing state-of-the-art framework, regions with convolutional neural networks (R-CNN), was proposed by Girshick \textit{et al.}~\cite{Girshick2015}. This framework consists of two steps during which (i) object areas are extracted as object proposals, and (ii) CNN recognition is performed. Those authors adopted selective search~\cite{Uijlings2013} as an object proposal approach and VGGNet for the CNN architecture. However, while they restricted the object detection and recognition tasks to fully connected CNN features, we believe that the features of the other layers should be more carefully evaluated in order to determine whether they could provide more accurate recognition and detection.

\section{Feature settings and representations}
In this paper, we evaluate two deep learning feature types. Figure \ref{fig:alex_vgg} shows the architectures of AlexNet~\cite{Krizhevsky2012} and VGGNet~\cite{Simonyan2015}. We believe that while the evaluation itself is very important, particular attention must be paid to tunings such as concatenation and feature transformation. Basically, deep learning architectures are based on their approaches. 

\textbf{Feature setting.} We begin by extracting the middle and deeper layers. Layers 3--7 of AlexNet and VGGNet are shown in Figure \ref{fig:alex_vgg}. Next, we extract each max-pooling layer (layers 3--5), and the last two fully connected layers (layers 6 and 7) in VGGNet. 

\textbf{Concatenation and transformation.} Next, we concatenate neighboring or one-step layers such as layer-3,4,5 and layer-3,5,7. In feature transformation, we simply apply PCA, which is set at 1,500 dimensions in this experiment. 

\textbf{Classifier.} In the next step, we apply deep learning features and SVM for object recognition. The parameters are based on DeCAF~\cite{Donahue2014}. 

\begin{figure*}[t]
\begin{center}
   \includegraphics[width=0.8\linewidth]{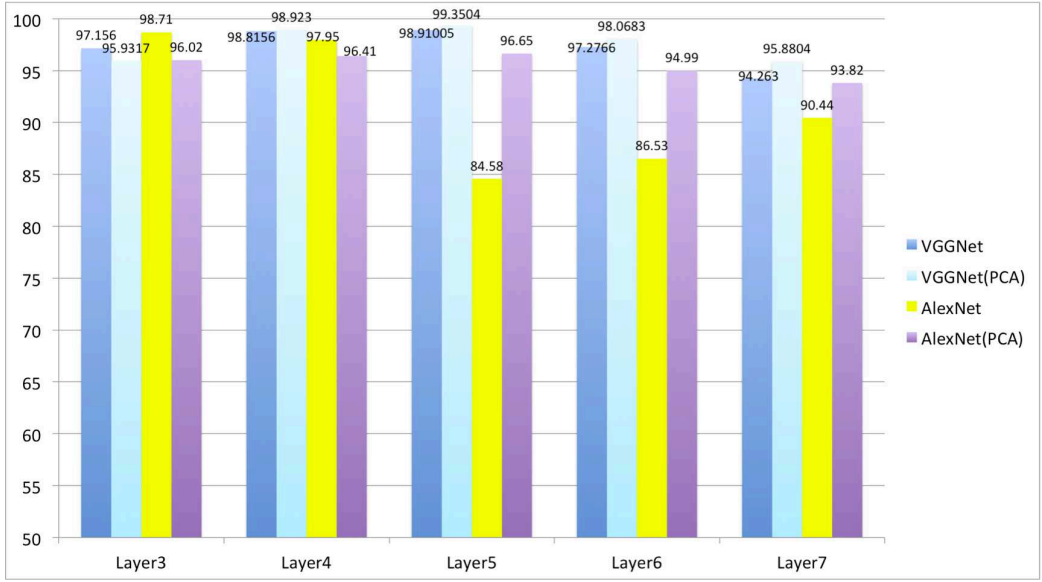}
\end{center}
   \caption{Comparison of AlexNet and VGGNet features on the Daimler Pedestrian Benchmark Dataset. CNN layers 3--7 are listed.}
\label{fig:daimler}
\end{figure*}

\begin{figure*}[t]
\begin{center}
   \includegraphics[width=0.8\linewidth]{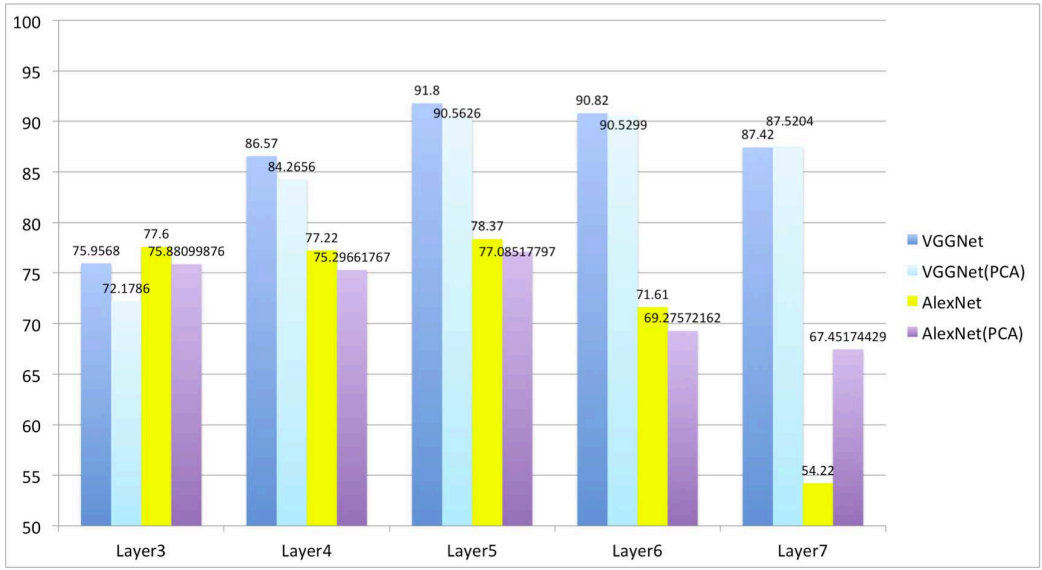}
\end{center}
   \caption{Comparison of AlexNet and VGGNet features on the Caltech 101 Dataset. CNN layers 3--7 are listed.}
\label{fig:caltech101}
\end{figure*}

\section{Experiments}
In this section, we discuss our experiments conducted using the Daimler pedestrian benchmark ~\cite{Munder2006} and Caltech 101 ~\cite{FeiFei2004} Datasets. Figure~\ref{fig:daimler} and ~\ref{fig:caltech101} show the results of our deep CNN feature evaluations on the Daimler and Caltech 101 datasets, respectively. The figures also show VGGNet, AlexNet, and their compressed features with PCA (VGGNet(PCA) and AlexNet(PCA)). 

In the Daimler dataset experiment, we found that the VGGNet(PCA) layers 5 and 4 showed the best performance rates at 99.35\% and 98.92\%, respectively. We also determined that PCA transforms low-dimensional features and feature vectors at better rates than the original features. The VGGNet layer 5 (98.91\%) and layer 4 (98.81\%) are, respectively, +0.44\% and +0.11\% improved with PCA. When AlexNet is used, layers 3 and 4 show top rates of 98.71\% and 97.95\%, respectively. As for VGGNet, layers 5 and 6 achieved the best results (91.8\%) on the Caltech 101 dataset. However, these results show significant layer 5 differences between VGGNet (91.8\%) and AlexNet (78.37\%).

From the above results, it can be seen that features obtained from fully connected layers do not always provide the highest performance rates during recognition and detection tasks, and that middle-layer features are more flexible for some tasks. We also found that fully connected layers or max-pooling layers located near fully connected layers tend to perform better in general object recognition tasks, such as the Caltech 101 dataset. 

The main difference between AlexNet and VGGNet is the architecture depth. Additionally, VGGNet assigns very small 3 $\times$ 3 convolutional kernels against the 7 $\times$ 7 (Conv 1), 5 $\times$ 5 (Conv 2), and 3 $\times$ 3 (others) kernels in AlexNet. The settings refrain the feature representation. 

The classification results of concatenated vectors are shown in Table~\ref{table:daimler} and ~\ref{table:caltech}. Here, it can be seen that concatenation of VGGNet layer-5,6,7 provides the highest levels of accuracy for both datasets. The rates are 99.38\% on the Daimler dataset and 92.00\% on the Caltech 101 dataset. For AlexNet, layer-3,4,5 and layer-3,5,7 achieved top performance rates on those datasets. The results show that combining features of the convolutional and fully connected layers provides better performance. It is especially noteworthy that VGGNet layer 5, which is near the fully connected layer, provides significantly high levels of feature extraction from an image patch.

\begin{table}[t]
\caption{Feature concatenation on the Daimler pedestrian benchmark dataset. The highest rate for each architecture is shown in bold.}
\begin{center}
\begin{tabular}{ccc}
\hline
Layer & VGGNet & AlexNet \\
\hline\hline
345 & 97.38 & \textbf{97.45} \\
456 & 98.73 & 96.98 \\
567 & \textbf{99.38} & 97.04 \\
357 & 95.96 & 97.26 \\
\hline
\end{tabular}
\end{center}
\label{table:daimler}
\end{table}

\begin{table}[t]
\caption{Feature concatenation on the Caltech 101 dataset. The highest rate for each architecture is shown in bold.}
\begin{center}
\begin{tabular}{ccc}
\hline
Layer & VGGNet & AlexNet \\
\hline\hline
345 & 78.13 & 77.95 \\
456 & 85.03 & 77.06 \\
567 & \textbf{92.00} & 77.38 \\
357 & 73.07 & \textbf{77.91} \\
\hline
\end{tabular}
\end{center}
\label{table:caltech}
\end{table}

\section{Conclusion}
In this paper, we evaluated two different of convolutional neural network (CNN) architectures AlexNet and VGGNet. The convolutional features from layers 3--7 were performed on the Daimler pedestrian benchmark and Caltech 101 datasets. We then attempted to implement feature concatenation and PCA transformation. Our experimental results show that the fully connected layers did not always perform better for recognition tasks. Additionally, the experiments using the Daimler and Caltech 101 datasets showed that layer 5 tends to provide the highest level of accuracy, and that feature concatenation of convolutional and fully connected layers improves recognition performance.


\begin{thebibliography}{10}\itemsep=-1pt

\bibitem{Donahue2014}
J.~Donahue, Y.~Jia, J.~Hoffman, N.~Zhang, E.~Tzeng, and T.~Darrell.
\newblock Decaf:a deep convolutional activation feature for generic visual
  recognition.
\newblock International Conference on Machine Learning (ICML), 2014.

\bibitem{FeiFei2004}
L.~Fei-Fei, R.~Fergus, and P.~Perona.
\newblock Learning generative visual models from few training examples: an
  incremental bayesian approach tested on 101 object categories.
\newblock IEEE Conference on Computer Vision and Pattern Recognition Workshop
  on Generative-Model Based Vision (CVPRW), 2004.

\bibitem{Fukushima1980}
K.~Fukushima.
\newblock Neocognitron: A self-organizing neural network model for a mechanism
  of pattern recognition unaffected by shift in position, 1980.

\bibitem{Girshick2015}
R.~Girshick, J.~Donahue, T.~Darrell, and J.~Malik.
\newblock Region-based convolutional networks for accurate object detection and
  segmentation.
\newblock IEEE Transactions on Pattern Analysis and Machine Intelligence
  (TPAMI), 2015.

\bibitem{HeCVPR2015}
K.~He and J.~Sun.
\newblock Convolutional neural networks at constrained time cost.
\newblock IEEE Conference on Computer Vision and Pattern Recognition (CVPR),
  2015.

\bibitem{Hearxiv2015}
K.~He, X.~Zhang, S.~Ren, and J.~Sun.
\newblock Delving deep into rectifiers: Surpassing human-level performance on
  imagenet classification.
\newblock arXiv:1502.01852, 2015.

\bibitem{Hinton2006}
G.~E. Hinton, S.~Osindero, and Y.-W. Teh.
\newblock A fast learning algorithm for deep belief nets.
\newblock Neural Computation, 2006.

\bibitem{Hinton2012}
G.~E. Hinton, N.~Srivastava, A.~Krizhevsky, I.~Sutskever, and R.~Salakhutdinov.
\newblock Improving neural networks by preventing co-adaptation of feature
  detectors.
\newblock CoRR, 2012.

\bibitem{Krizhevsky2012}
A.~Krizhevsky, I.~Sutskever, and G.~E. Hinton.
\newblock Imagenet classification with deep convolutional neural networks.
\newblock NIPS, 2012.

\bibitem{LeCun1989}
Y.~LeCun, B.~Boser, J.~S. Denker, D.~Henderson, R.~E. Howard, W.~Hubbard, and
  L.~D. Jackel.
\newblock Backpropagation applied to handwritten zip code recognition.
\newblock Neural Computation, 1989.

\bibitem{Munder2006}
S.~Munder and D.~M. Gavrila.
\newblock Daimler mono pedestrian classification benchmark dataset: An
  experimental study on pedestrian classification, 2006.

\bibitem{Nair2010}
V.~Nair and G.~E. Hinton.
\newblock Rectified linear units improve restricted boltzmann machines.
\newblock International Conference on Machine Learning (ICML), 2010.

\bibitem{Russakovsky2014}
O.~Russakovsky, J.~Deng, H.~Su, J.~Krause, S.~Satheesh, S.~Ma, Z.~Huang,
  A.~Karpathy, A.~Khosla, M.~Bernstein, A.~C. Berg, and L.~Fei-Fei.
\newblock Imagenet large scale visual recognition challenge.
\newblock arXiv:1409.0575, 2014.

\bibitem{Ioffe2015}
I.~S. and C.~Szegedy.
\newblock Batch normalization: Accelerating deep network training by reducing
  internal covariate shift.
\newblock arXiv:1502.03167, 2015.

\bibitem{Schmidhuber2015}
J.~Schmidhuber.
\newblock Deep learning in neural networks: An overveiw, 2015.

\bibitem{Simonyan2015}
K.~Simonyan and A.~Zisserman.
\newblock Very deep convolutional networks for large-scale image recognition.
\newblock International Conference on Learning Representation (ICLR), 2015.

\bibitem{Szegedy2015}
C.~Szegedy, W.~Liu, Y.~Jia, P.~Sermanet, S.~Reed, D.~Anguelov, D.~Erhan,
  V.~Vanhoucke, and A.~Rabinovich.
\newblock Going deeper with convolutions.
\newblock IEEE Conference on Computer Vision and Pattern Recognition (CVPR),
  2015.

\bibitem{Uijlings2013}
J.~R.~R. Uijlings, K.~E.~A. van~de Sande, T.~Gevers, and A.~W.~M. Smeulders.
\newblock Selective search for object recognition, 2013.

\end{thebibliography}
\end{document}